%% file: main.tex
\documentclass[acmsmall, nonacm]{acmart}

\input{macros}

\usepackage[autostyle, english = american]{csquotes}
\MakeOuterQuote{"}
\usepackage{subcaption}

\AtBeginDocument{%
  \providecommand\BibTeX{{%
    \normalfont B\kern-0.5em{\scshape i\kern-0.25em b}\kern-0.8em\TeX}}}

\setcopyright{acmcopyright}
\copyrightyear{2018}
\acmYear{2018}
\acmDOI{XXXXXXX.XXXXXXX}

\acmConference[Conference acronym 'XX]{Make sure to enter the correct
  conference title from your rights confirmation emai}{June 03--05,
  2018}{Woodstock, NY}
\acmBooktitle{Woodstock '18: ACM Symposium on Neural Gaze Detection,
 June 03--05, 2018, Woodstock, NY} 
\acmPrice{15.00}
\acmISBN{978-1-4503-XXXX-X/18/06}

\begin{document}

\title{Rethinking Streaming Machine Learning Evaluation}

\author{Shreya Shankar}
\authornote{Both authors contributed equally to this work.}
\affiliation{%
  \institution{University of California, Berkeley}
  \country{USA}
}
\email{shreyashankar@berkeley.edu}

\author{Bernease Herman}
\authornotemark[1]
\affiliation{%
  \institution{University of Washington}
  \country{USA}
}
\email{bernease@uw.edu}

\author{Aditya G. Parameswaran}
\affiliation{%
  \institution{University of California, Berkeley}
  \country{USA}
}
\email{adityagp@berkeley.edu}



\begin{abstract}
    While most work on evaluating machine learning (ML) models focuses on computing accuracy on {\em batches} of data, tracking accuracy alone in a \emph{streaming} setting (i.e., unbounded, timestamp-ordered datasets) fails to appropriately identify when models are performing unexpectedly. In this position paper, we discuss how the nature of streaming ML problems introduces new real-world challenges (e.g., delayed arrival of labels) and recommend additional metrics to assess streaming ML performance. 
\end{abstract}

\maketitle

\section{Introduction}
\label{sec:intro}
\input{intro}

\section{Why Can't We Just Track Accuracy?}
\label{sec:problems}
\input{problems}

\section{Towards Better Streaming Metrics}
\label{sec:towards}
\input{towards}

\section{Conclusion}
\input{conclusion}

\bibliographystyle{ACM-Reference-Format}
\bibliography{references}

\end{document}

%% file: macros.tex
\newcommand{\topic}[1]{\vspace{-3.5pt}\smallskip \smallskip \noindent{\bf #1.}}

%% file: intro.tex

As machine learning (ML) models are embedded in software systems, a new research question emerges: how do we evaluate ML performance on \emph{streams} of data (i.e., unbounded, timestamp-ordered datasets)? When developing models, we typically evaluate them by partitioning a dataset into fixed training, validation, and test sets; for each partition, we compute an aggregate metric over all data points. While this \emph{batch} evaluation paradigm is reasonable for many temporally-agnostic tasks, it does not provide a reliable assessment for ML performance over time.

It is now well-known that the nature of inputs to deployed models may change over time, motivating recent benchmarks that involve datasets drawn from different distributions~\cite{wilds, breeds, Hendrycks2021TheMF, lin2021the}. While these benchmarks have led to new learning algorithms~\citep{dabs}, data augmentation techniques~\cite{noiseaug}, and evaluation methods~\cite{Kaplun2022DeconstructingDA} to adapt models to and understand their performance on new datasets, these papers mainly evaluate in batch, wherein data points are weighted equally or by subpopulation, but not temporally.

In the streaming setting, we desire some \emph{recency bias}: recent predictions should be prioritized more than earlier predictions. Existing work on understanding how ML performance changes over streams of data usually evaluates accuracy in batch over large windows of time (e.g., years)~\citep{ease_ml, lin2021the}. In this case, evaluation involves a (1) metric function and (2) window (i.e., defined based on a fixed number of data points or fixed duration fed into the function). Post-deployment evaluation typically involves at least one of two types of windows, which define (possibly overlapping) partitions of the data:

\begin{itemize}
    \item \textbf{Cumulative window}, where all data is included in a single partition. This often leads to evaluation that treats distant data points equally to more recent ones.
    \item \textbf{Sliding window}, where data from the most recent $n$ data points or units of time (e.g., days) comprise each partition. In either case, this is a fixed window.
\end{itemize}


Determining a window size for an ML task is not the only challenge in evaluating ML performance in a streaming setting. The nature of the ML task may change over time (e.g., transition from a balanced binary classification problem to a highly imbalanced problem). As a result, accuracy may not be perfectly aligned with organizational needs. 

In this position paper, we argue that evaluating only accuracy over sliding windows is not actionable enough for practitioners.  Using real-world data, in Section~\ref{sec:problems}, we demonstrate the impacts of three phenomena --- (1) varying window sizes, (2) representation differences, and (3) delayed and incomplete labels --- on accuracy. Finally, in Section~\ref{sec:towards}, we recommend additional metrics to track in the streaming setting, where the nature of the ML task might change over time.

%% file: problems.tex
To illustrate how accuracy on a fixed window size doesn't work, we first describe an example ML task and data.

\subsection{Preliminaries}

\topic{ML Task} Using data from the New York City Taxi and Limousine Coalition~\cite{nyc-taxi}, our ML task involves predicting whether a rider will give their driver a tip (defined as > 10\% of the fare). Our task therefore involves binary classification, where predictions are probabilities (i.e., are floats between 0 and 1). 

\topic{Data} We limit ourselves to data from 2020. Each row in the dataset corresponds to a single cab ride, with 17 attributes. We generate features from these attributes and encode them such that all are treated as numerical variables. There are approximately a few million entries (rows) per month.

\topic{ML Model} We use a \texttt{scikit-learn} random forest classifier for simplicity. The ML model is evaluated on accuracy, or the fraction of correct predictions, when the prediction is rounded to the nearest integer. Training accuracy corresponds to accuracy on the training set (i.e., batch accuracy), and live accuracy corresponds to accuracy measured post-deployment.

\topic{Window Size} We train and validate a model on January 2020 data and "deploy" on February 1, 2020. For the streaming evaluation metric, we evaluate accuracy on a 7-day sliding window. 

\subsection{Impacts of Real-World Phenomena on Accuracy}

The goal of an evaluation metric should be to assess whether models are performing as we expect them to. Accuracy alone does not address this and suffers from a number of failure modes, which we subsequently describe.

\begin{figure}
    \centering
    \includegraphics[width=0.9\linewidth]{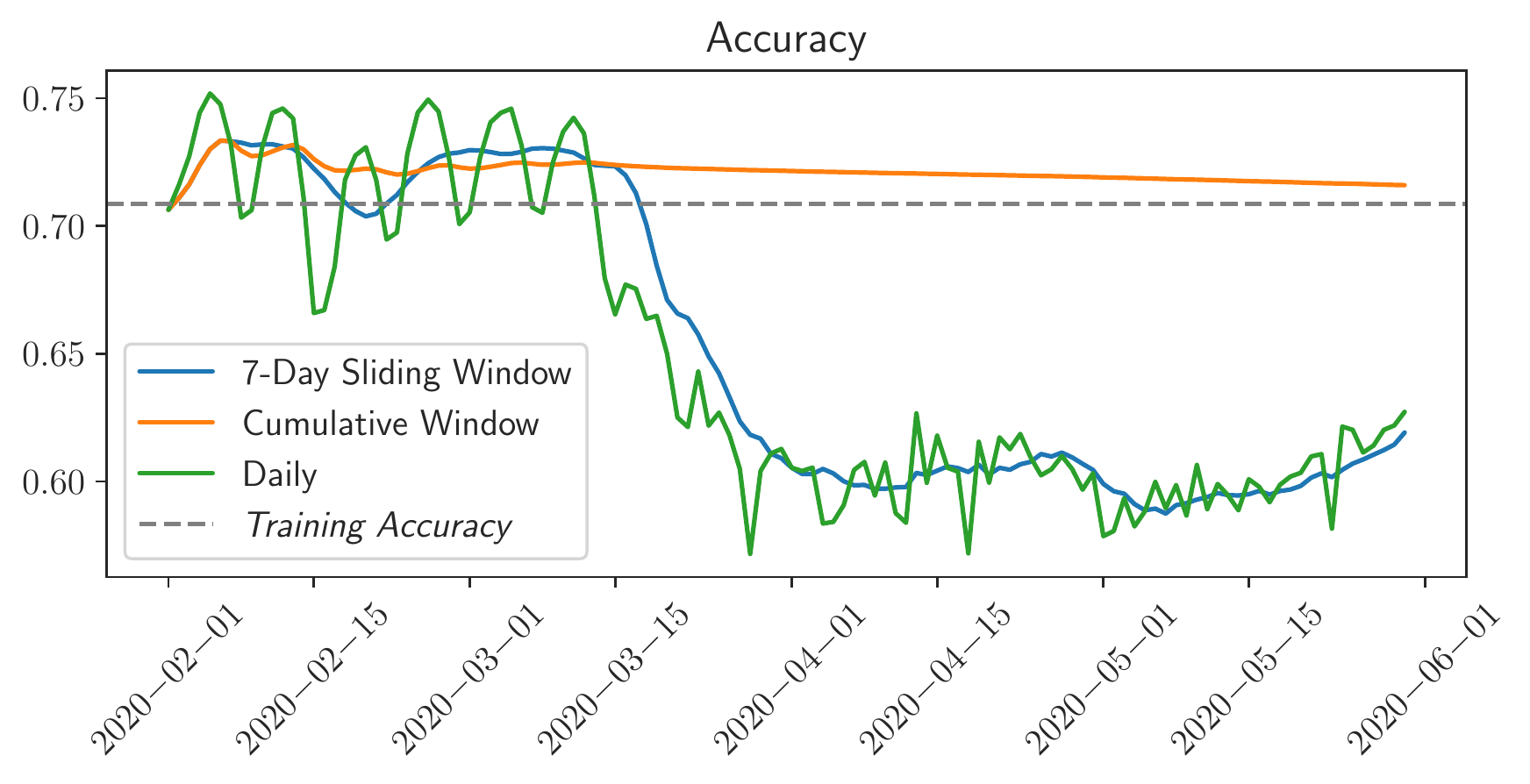}
    \caption{Accuracy for a model evaluated over cumulative, 7-day sliding, and daily windows.}
    \label{fig:accuracies}
\end{figure}

\begin{figure*}[t!]
    \centering
    \begin{subfigure}[t]{0.48\linewidth}
        \centering
        \includegraphics[width=\linewidth]{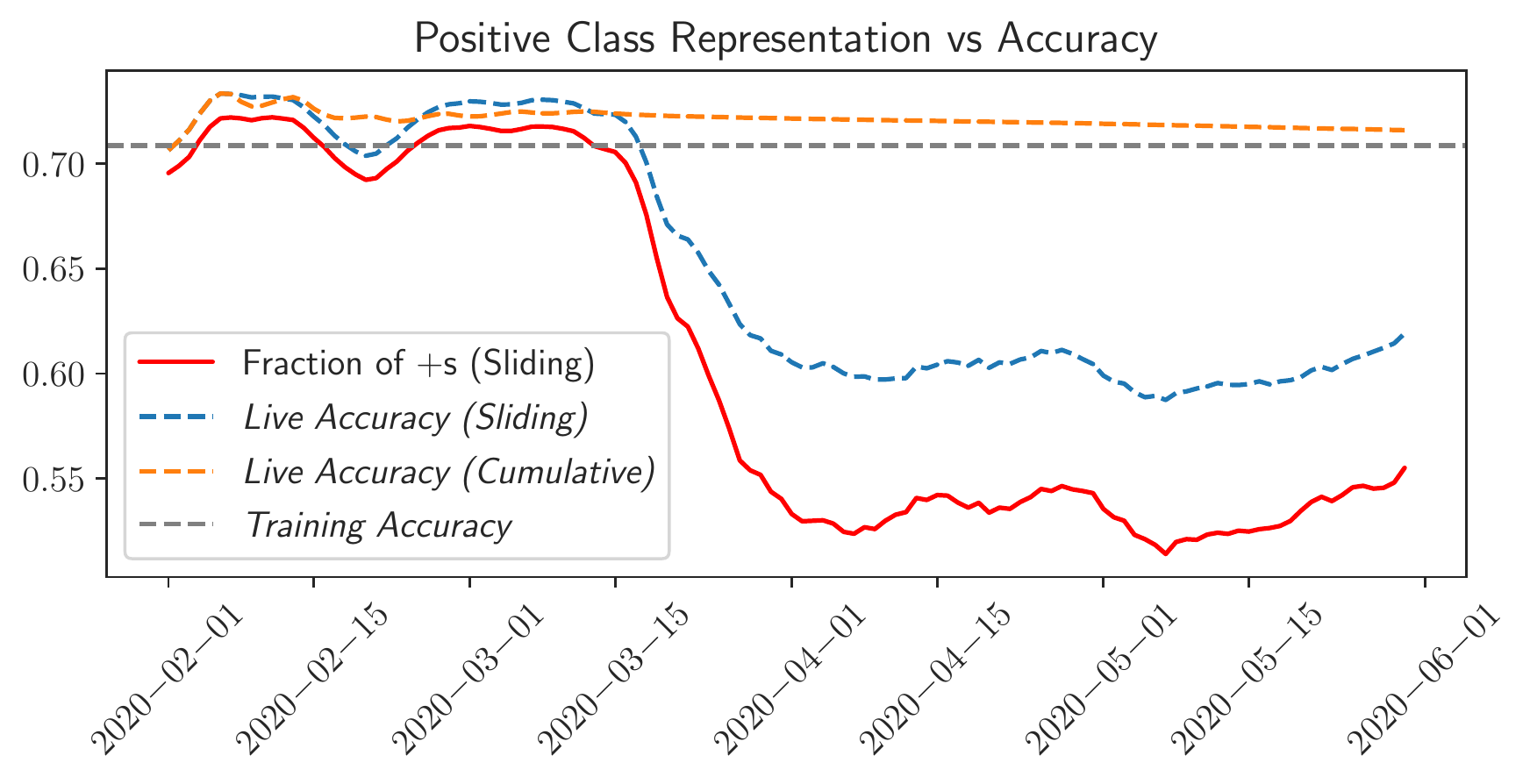}
        \caption{Fraction of positive examples in the 7-day sliding window (red line). The positive class representation drops significantly in mid-March, presumably due to the onset of the COVID-19 pandemic. For reference, model accuracy is depicted by dashed lines.}
        \label{fig:classreps}
    \end{subfigure}
    \hspace{0.5em}
     \begin{subfigure}[t]{0.48\linewidth}
        \centering
        \includegraphics[width=\linewidth]{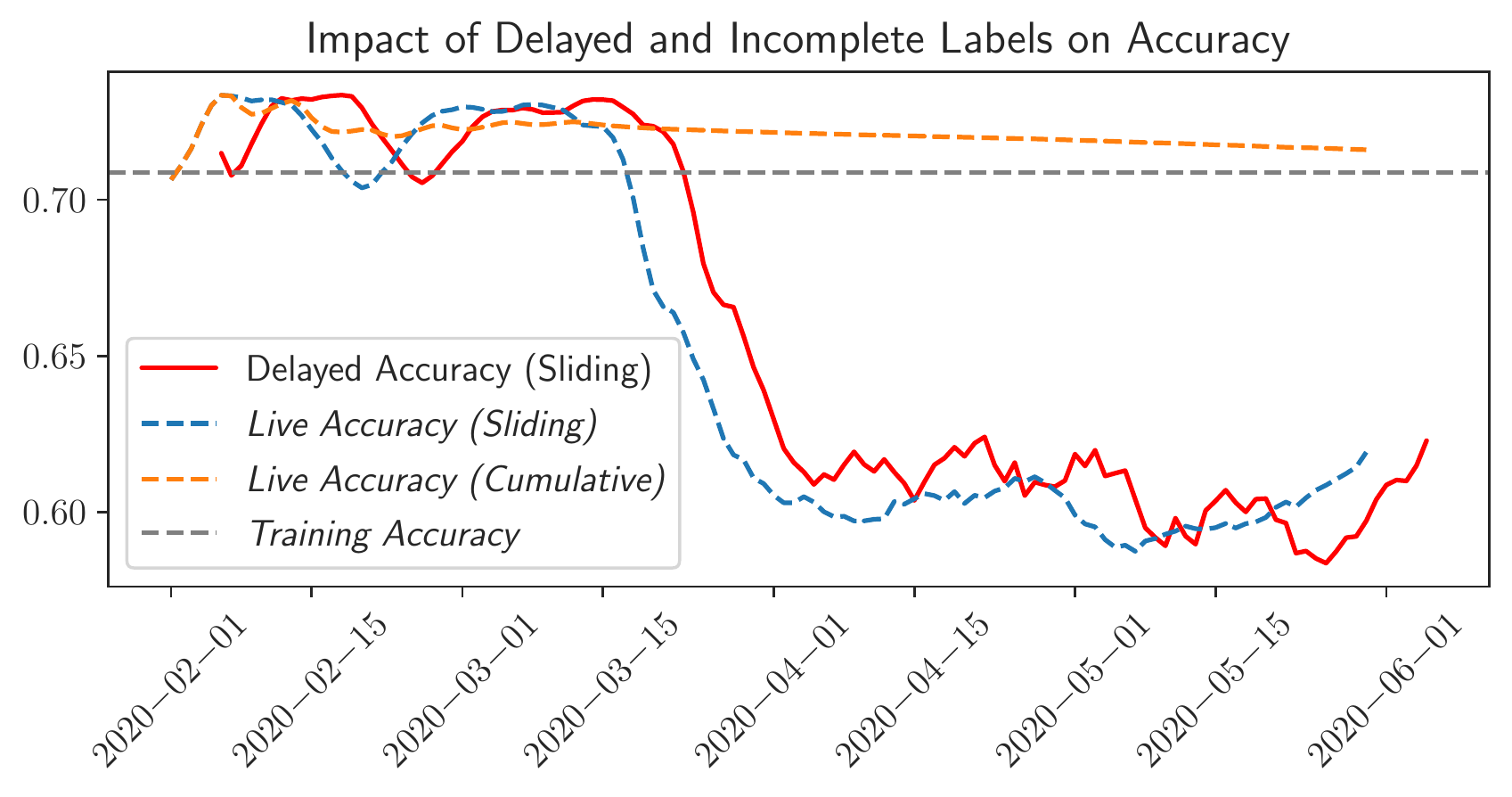}
        \caption{Accuracy when when only 10\% of predictions are labeled and arrive after a delay (red line). Delays are randomly sampled as $d \sim \mathit{Exp}\left(\lambda = 7\right)$, and the new timestamp becomes the original + $d$ days. Delayed and incomplete labels cause accuracy to look different than expected (blue dashed line).}
        \label{fig:delays}
    \end{subfigure}%
    \caption{Accuracy over time with respect to real-world phenomena.}
\end{figure*}

\subsubsection{Varying Window Sizes} First, adopting a fixed widow size can be problematic, since metric values are sensitive to window size. This effect is shown in Figure \ref{fig:accuracies} where for the same choice of metric across different window sizes (e.g., cumulative vs sliding) results in significantly different outcomes. The cumulative window barely registers the accuracy drop. Second, although a 7-day sliding window has a smoother performance than the daily accuracy (i.e., 1-day sliding window), identifying drastic drops in accuracy can be important for model evaluation in particular settings (e.g., medical settings or healthcare ML deployments). However, in other settings, the smoothed sliding window with the larger size may highlight global patterns (e.g., trends over weeks as opposed to weekdays/weekends). A pain point for practitioners is that they must think carefully about how evaluation window size impacts reported performance.

\subsubsection{Representation Differences}
\label{sec:repdiffs}
Class ratios across windows may not be the same (e.g., the fraction of positives in one window may be very different from the fraction of positives in another window). As a result, live accuracy can vary wildly even when the expected quality of the model did not degrade. For instance, in Figure~\ref{fig:classreps}, the positive class representation curve mirrors the shape of the live accuracy curve for the same sliding window, indicating that the tip prediction model performance decrease might be attributed to the fact that fewer riders are tipping their drivers. However, without looking at the positive class representation curve, we might conclude that there is a change in \emph{how} riders tip their drivers --- and thus retrain our models blindly on recent data without accounting for representation differences (e.g., not balancing classes). Moreover, as class representations change, different metric functions might need to be computed --- e.g., F1 score might be a better choice than accuracy --- to tell us when models don't have the predictive power that we expect.

\subsubsection{Delayed and Incomplete Labels}
\label{sec:delayedlabels}
Due to the nature of many tasks and software systems, labels may come in after some delay, or not at all. This delay makes it almost impossible to compute, in real-time, exact accuracy for predictions made in a specific window of time. We ran a small experiment to demonstrate the impact of delayed and incomplete labels on accuracy by sampling some delay $d \sim \mathit{Exp}\left(\lambda = 7\right)$ for each row and adding $d$ days to the original timestamp in that row. We labeled only 10\% of the predictions made. Figure~\ref{fig:delays} shows that delayed and incomplete labels can cause the accuracy curve (red) to look different than the true accuracy curve (blue). If we can't even observe the true accuracy curve, how are we supposed to know when to retrain models?

%% file: towards.tex
As demonstrated in Section~\ref{sec:problems}, changes in accuracy can be misaligned with global ML performance due to varying window sizes, class representation differences, or label delays. As a result, it can be challenging to understand  \emph{when} and \emph{how} models aren't performing as expected. In this section, we motivate properties of good streaming ML metrics and recommend additional metrics for practitioners to track.


\subsection{Desired Properties}

To address some of the issues outlined in Section~\ref{sec:problems}, an ideal streaming ML metric should have the following properties:
\begin{itemize}
    \item Incorporates recency bias (i.e., sensitivity to local failures)
    \item Utilizes ground truth labels (e.g., not solely statistical measures of features) if available
    \item Reflects values specific to the organizational values, task, and setting in which the metric will be applied
\end{itemize}

While evaluating accuracy over sliding windows addresses recency bias and involves ground-truth labels, it is difficult to understand whether models are performing unexpectedly and thus deviating from organizational values. For example, in the streaming setting, if models are seeing significantly fewer positive labels, then we may need to track precision and recall. Or, if models are receiving significantly more data points from an underrepresented subpopulation in the training set, we should expect accuracy to decrease. What metrics tell us when models are performing \emph{unexpectedly}?

\subsection{Evaluating Unexpected Performance Drops}

As discussed in Section~\ref{sec:repdiffs}, changes in class representations can motivate us to care about different metrics --- for example, if the class imbalance is very high, we should track precision and recall in addition to accuracy. Thus, we recommend tracking class representations (i.e., fraction of examples in each class) over sliding windows to alert further investigation of the model. 

To address the delayed or incomplete label phenomenon as described in Section~\ref{sec:delayedlabels}, one idea is to use importance weighting (IW)~\cite{sugiyama}. At a high level, we identify subgroups 
based on input features or combinations thereof, 
determine the training set accuracy 
for each subgroup, and weight these accuracies 
based on the number of points in each subgroup 
in the live data. Incorporating IW estimates into streaming ML metrics is unexplored in both researcher and practitioner communities. 

\begin{figure*}[t!]
    \centering
    \begin{subfigure}[t]{0.4\linewidth}
        \centering
        \includegraphics[width=\linewidth]{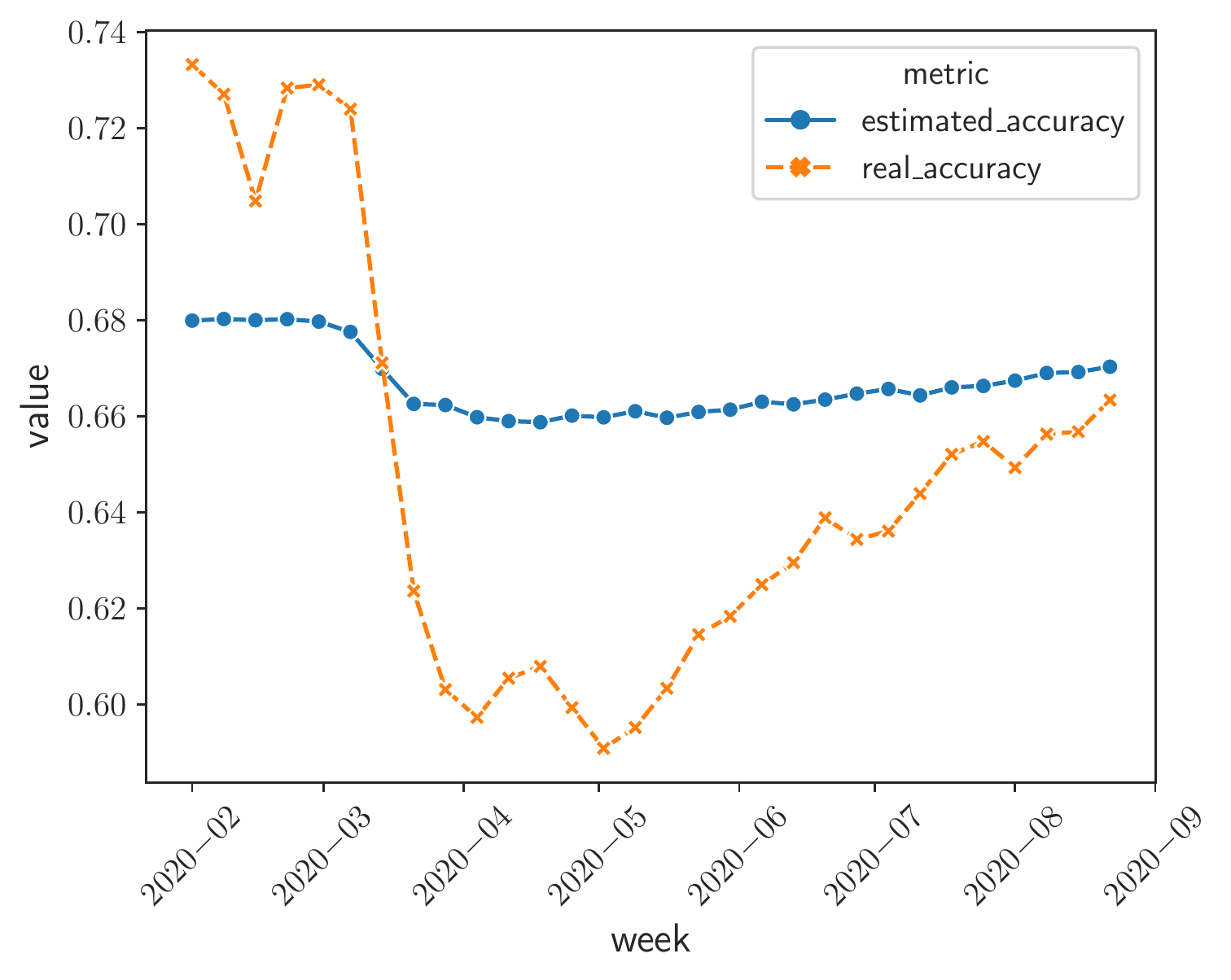}
        \caption{Importance-weighted accuracy estimate vs real accuracy (computed weekly). The divergence in metrics indicates a concept shift --- around the onset of the COVID-19 pandemic.}
        \label{fig:iwaccuracies}
    \end{subfigure}
    \hspace{0.5em}
     \begin{subfigure}[t]{0.57\linewidth}
        \centering
        \includegraphics[width=\linewidth]{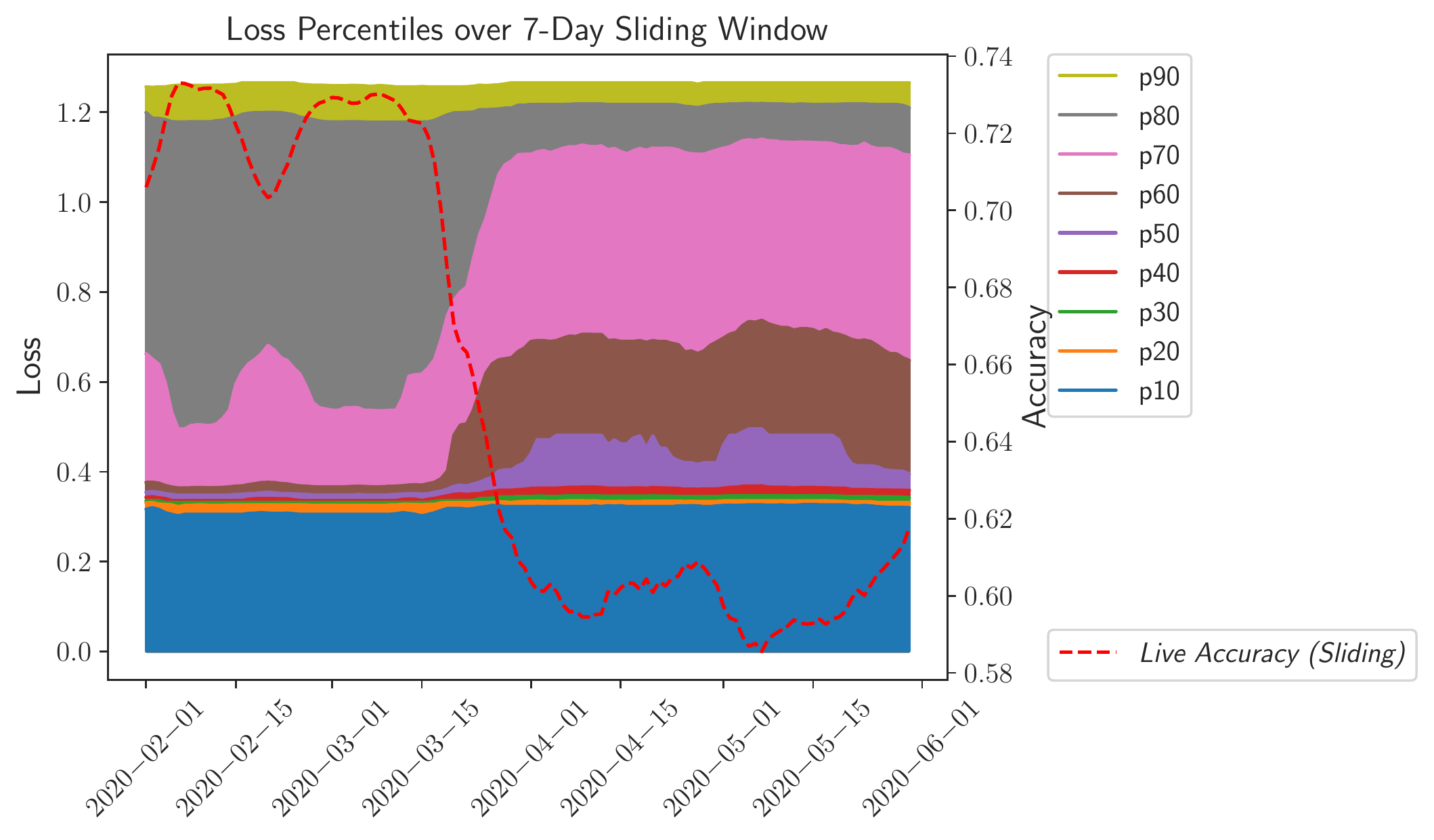}
        \caption{Loss percentiles for 7-day sliding windows over time. The live accuracy is depicted by the dashed red line. The large jump in loss percentile values coincides with the large accuracy drop.}
        \label{fig:percentiles}
        \end{subfigure}%
    \caption{Additional metrics to track.}
\end{figure*}

\topic{IW Difference Metrics} To start, we can track the difference between the IW estimate and the practical accuracy over sliding windows. Figure~\ref{fig:iwaccuracies} shows IW estimates based on pickup location and real accuracy, tracked weekly. The large difference between the real and estimated accuracies indicates a concept shift, or a change in how riders are tipping drivers, motivating a retrain of the model. In our case, the pickup location neighborhood is a simple subgrouping strategy. We may want our subgroups to be more or less fine-grained based on how much error we are willing to tolerate in the IW estimate~\cite{shankar2021observability}. 

\topic{Tracking Model Loss} IW estimates might tell us when to retrain models, but how to retrain models is still an open question. What subgroups are underrepresented? An interesting idea is to track the distribution of the model's loss over time. Loss percentiles for sliding windows is shown in Figure~\ref{fig:percentiles}. The most significant loss increase around the onset of the live accuracy drop is in the 70th percentile loss group. We could upweight these data points during retraining, or do some weighted random subsampling to construct a set of data points for online fine-tuning.

%% file: conclusion.tex
In this position paper, we advocate for tracking additional metrics for streaming ML tasks. In addition to accuracy, we propose for practitioners to track several metrics over sliding windows. To deal with high variability in metric values over time, employ a range of different window sizes. To address the fact that class representations can vary over time, track the fraction of examples in each class. To handle label delays, track importance-weighted (IW) estimates of the metric. When labels arrive, track differences between the IW estimate and the real metric value and the distribution of model loss (e.g., percentiles).

We urge practitioners to think critically about the metrics they want to track --- metrics that consider recency bias, involve some ground truth labels, and ultimately help us understand when and how models are failing in the wild after deployment.

